\pdfoutput=1

\documentclass[11pt]{article}

\usepackage{acl}

\usepackage{times}
\usepackage{latexsym}
\usepackage{amsmath} 
\usepackage{booktabs} 
\usepackage{multirow}
\usepackage{colortbl}

\usepackage[T1]{fontenc}

\usepackage[utf8]{inputenc}

\usepackage{microtype}

\usepackage{inconsolata}

\usepackage{graphicx}

\usepackage{tcolorbox}
\tcbuselibrary{breakable} 

\title{Teaching Vision-Language Models to Ask: Resolving  \\ Ambiguity in Visual Questions}

\author{
    Pu Jian\textsuperscript{1,2}, \ 
    Donglei Yu\textsuperscript{1,2},\ 
    Wen Yang\textsuperscript{1,2}, \
    Shuo Ren\textsuperscript{1}, \
    Jiajun Zhang\textsuperscript{1,2,3}\thanks{\ \ Corresponding Author}   \\
    \textsuperscript{1}Institute of Automation, Chinese Academy of Sciences\\
    \textsuperscript{2}School of Artificial Intelligence, University of Chinese Academy of Sciences\\
    \textsuperscript{3}Wuhan AI Research\\
    \texttt{\{jianpu2023, yudonglei2021, yangwen2023, shuo.ren\}@ia.ac.cn}\\
    \texttt{jjzhang@nlpr.ia.ac.cn} \\
}

\begin{document}
\maketitle
\begin{abstract}
In visual question answering (VQA) context, users often pose ambiguous questions to visual language models (VLMs) due to varying expression habits. Existing research addresses such ambiguities primarily by rephrasing questions. These approaches neglect the inherently interactive nature of user interactions with VLMs, where ambiguities can be clarified through user feedback. However, research on interactive clarification faces two major challenges: (1) Benchmarks are absent to assess VLMs' capacity for resolving ambiguities through interaction; (2) VLMs are trained to prefer answering rather than asking, preventing them from seeking clarification. To overcome these challenges, we introduce \textbf{ClearVQA} benchmark\footnote{The proposed ClearVQA dataset and related codes can be found at https://github.com/jian0805/ClearVQA}, which targets three common categories of ambiguity in VQA context, and encompasses various VQA scenarios.
Furthermore, we propose an automated pipeline to generate ambiguity-clarification question pairs. Experimental results demonstrate that training based on the automated generated data enables VLMs to ask reasonable clarification questions, thereby generating more accurate and specific answers based on user feedback.

\end{abstract}

\section{Introduction}

\begin{figure}[t]
  \centering
  \includegraphics[width=0.45\textwidth]{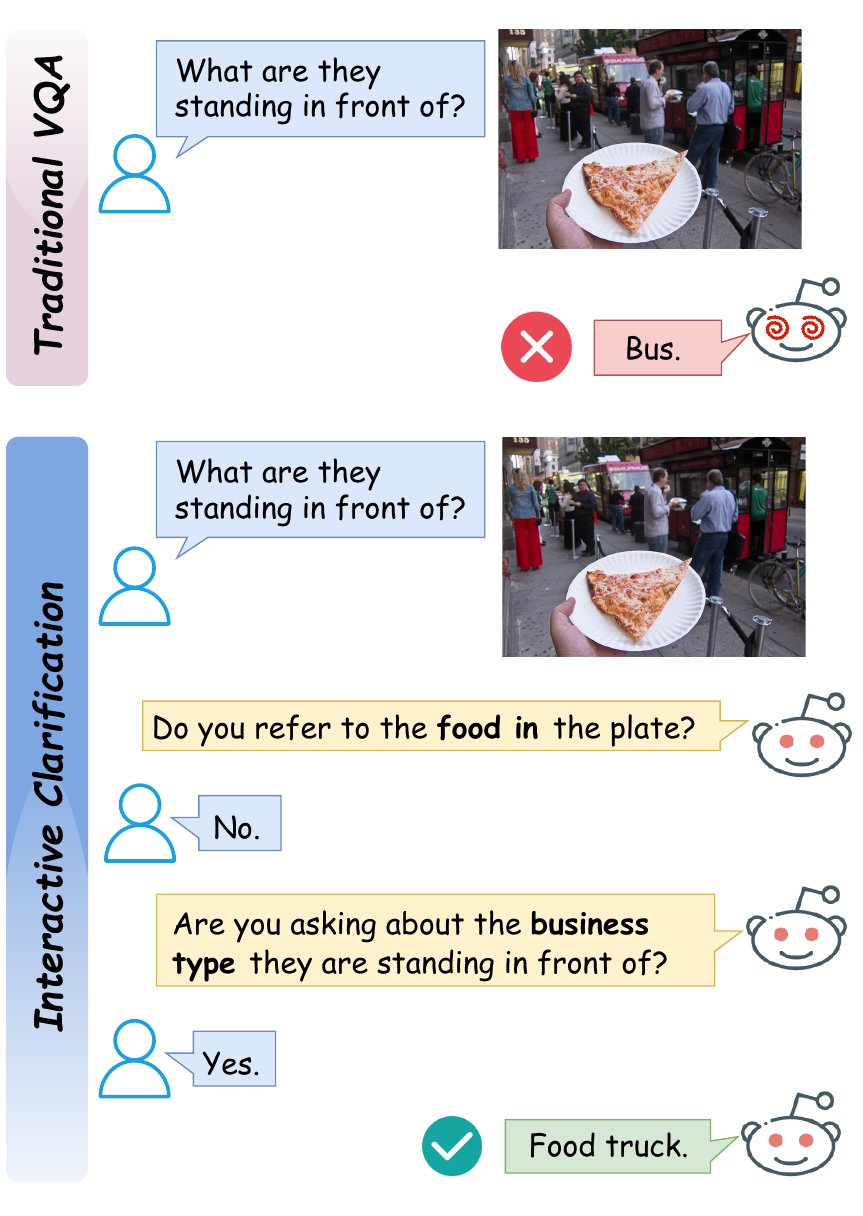}
  \caption{In the traditional VQA context, ambiguous questions may confuse VLMs and lead to undesired answers. In such cases, we emphasize that VLMs should pose clarification questions and generate desired answers based on user feedback.}
  \label{fig:clearvqa}
\end{figure}

The visual question answering (VQA) task aims to provide a natural language answer to a question based on a given image \cite{antol2015vqa, gao2022transform}. Recent studies in VQA field \cite{stengel2023did, prasad2023rephrase} highlight that users often pose ambiguous visual questions due to differences in language proficiency and expression habits \cite{ye2025genericempathypersonalizedemotional, wu2024macaroon}. For example, the question in Figure~\ref{fig:clearvqa} could be asking about the type of "\texttt{vehicle}" or "\texttt{business}" near the crowd, or it might refer to the "\texttt{pizza}" in the foreground. Since the prevailing assumption of VQA task is to return a single answer, when directly addressing ambiguous questions, even state-of-the-art large visual-language models (VLMs) \cite{liu2024improved} that excel in visual understanding and cross-modal reasoning may generate undesired responses.

Several studies have investigated how VLMs address ambiguous instructions or questions. \citet{pezzelle2023dealing} emphasize that the ambiguity of instructions is a significant source of errors in visual-language tasks. One feasible solution to such ambiguity is additional pre-training to align ambiguous text with images better, enabling VLMs to interpret such questions in a more human-like manner \cite{sun2024generative, lu2023lyrics}. However, this approach leads to significant computational costs. \citet{stengel2023did} introduce a question generation model that rephrases ambiguous visual questions based on images to reduce ambiguity. Similarly, \citet{prasad2023rephrase} leverage well-trained VLMs to extract visual details omitted from the questions and employ a large language model (LLM) to rephrase ambiguous questions in a zero-shot manner, thus improving VQA accuracy. 

Despite their contributions, these studies focus solely on inferring possible user intents to address ambiguity. They overlook the interactive nature of real-world VLM applications, where ambiguity can be resolved more effectively through user feedback. By engaging in an interactive clarification, VLMs can generate more accurate and specific answers \cite{naszadi2023aligning}. However, adopting this interactive approach faces two major challenges: 1) How to comprehensively evaluate the ability of VLMs to eliminate ambiguity through interactive approaches and accurately quantify this capability using appropriate metrics; 2) Such interactive approach requires VLMs to pose clarification questions, but most VLMs are primarily trained to answering rather than asking. 

\begin{figure*}[t]
  \centering
  \includegraphics[width=0.99\textwidth]{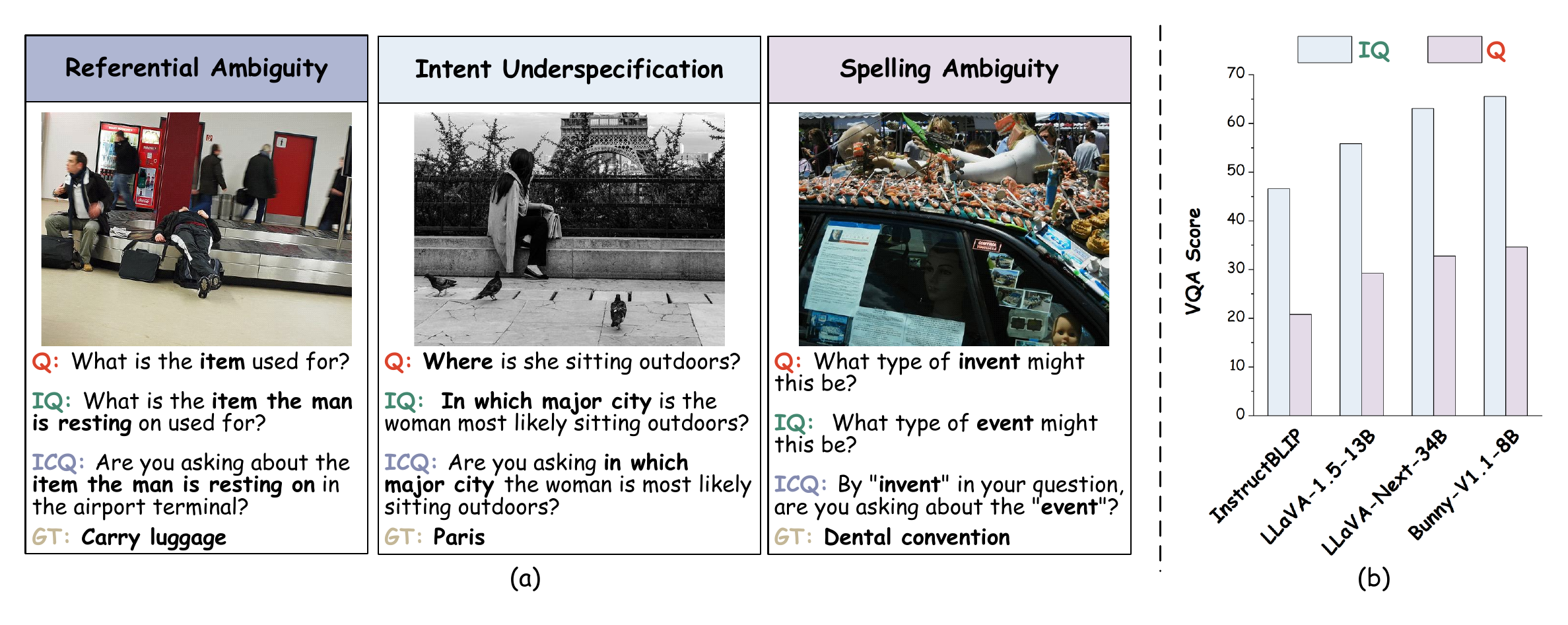}
  \caption{The ambiguity in visual questions emphasized in the ClearVQA Benchmark. (a) Ambiguity in ClearVQA is categorized into three types: referential ambiguity, intent underspecification, and spelling ambiguity. Q: question. IQ stands for the user’s intended question. ICQ represents the ideal clarification question. GT stands for the ground truth answer. (b) Experimental results on the test set show that, compared to explicitly intended IQ, existing VLMs struggle to handle the corresponding ambiguous question, leading to a significant drop in VQA accuracy.}
  \label{fig:ambiguity_category}
\end{figure*}

To address the first challenge, we introduce ClearVQA benchmark. ClearVQA emphasizes that for ambiguous visual questions, VLMs should ask clarification questions based on given images, and generate desired answers in an interactive dialogue using user feedback. ClearVQA focuses on three common categories of ambiguity observed in VQA datasets and real-world scenes \cite{bhattacharya2019does, prasad2023rephrase, pezzelle2023dealing}: {\textit{referential ambiguity}}, {\textit{intent underspecification}}, and {\textit{spelling ambiguity}}, as shown in Figure~\ref{fig:ambiguity_category}. Additionally, ClearVQA spans various VQA scenarios, including visual understanding, cross-modal reasoning, fine-grained knowledge, and scene text. 



To address the second challenge, we propose an automatic pipeline for generating ambiguity-clarification question pairs. This automated pipeline makes it possible to conduct specialized training for interactive clarification capability. Experimental results demonstrate that training based on automatically generated data enables open-source VLMs to effectively handle ambiguous visual questions. It is worth noting that our work marks an initial step in the multimodal field toward resolving ambiguity through interactive methods.



\section{ClearVQA Benchmark}
\label{sec:clearVQA benchmark}


\subsection{Task Formulation}
\label{task_formulation}
The primary setup for VQA requires VLMs to learn a function $f: X \mapsto \mathcal{A}$, where $\boldsymbol{x} = \begin{pmatrix}\boldsymbol{v},\boldsymbol{q}\end{pmatrix} \in X$ represents a question-image pair \cite{dancette2023improving}. However, in real-world scenes, the input questions may exhibit ambiguity due to variations in users' language abilities and expression habits. In such cases, we expect VLMs to ask clarification questions and, based on user feedback, generate more accurate and specific answers. Therefore, in the interactive clarification context, VLMs need to learn a function $h:\mathcal{X}\mapsto\mathcal{A}\cup\mathcal{C}$, thus the input to VLMs becomes
\begin{equation}
\boldsymbol{\sigma}_n = (\boldsymbol{x},\boldsymbol{c}_1, \boldsymbol{\xi}_1, \boldsymbol{c}_2, \boldsymbol{\xi}_2, \cdots, \boldsymbol{c}_n,\boldsymbol{\xi}_n) \in \mathcal{X}, 
\end{equation}
with $\boldsymbol{c}_i$ and $\boldsymbol{\xi}_i, i\in \{1,\cdots,n\}$ representing the clarification question posed by the VLMs and the user's feedback at the $i$-th turn, respectively, and $\boldsymbol{\sigma}_0 = \boldsymbol{x}$. Additionally, the output space is expanded to include the option to pose clarification questions $\boldsymbol{c} \in \mathcal{C}$. The function $h$ can be decomposed into three functions: the VQA function $f: \mathcal{X} \mapsto \mathcal{A}$, the clarification question generation function $g: \mathcal{X} \mapsto \mathcal{C}$, and the ambiguity detection function $\phi: \mathcal{X} \mapsto \{0,1\}$. In the context of multi-turn interactive clarification, the output at the $i$-th turn, $i \in \{1,2,3, \dots\}$ is given by
\begin{equation}
\begin{aligned}
y_{i} =& h(\boldsymbol{\sigma}_i)  = 
\begin{cases}
f(\boldsymbol{\sigma}_i) & \text{if}  \phi(\boldsymbol{\sigma}_i) = 1 \\
g(\boldsymbol{\sigma}_i) & \text{if}  \phi(\boldsymbol{\sigma}i) = 0
\end{cases}.
\end{aligned}
\end{equation}
Ideally, the iteration terminates if the VLMs output an answer instead of a clarification question.

\begin{figure*}[t]
  \centering
  \includegraphics[width=0.999\textwidth]{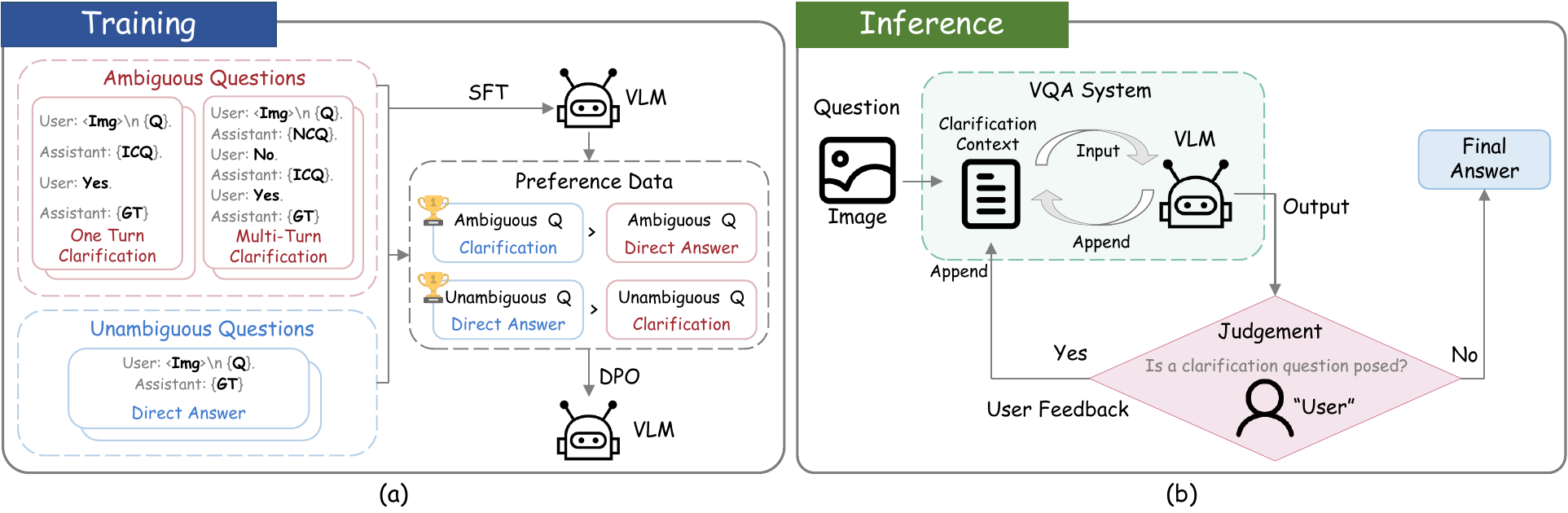}
  \caption{Training process (a) for enabling interactive clarification capabilities and inference process (b). ICQ represents the ideal clarification question. GT stands for the ground truth answer. Q stands for question. NCQ is the clarification question that fails to reflect the true intent of "\texttt{user}".}
  \label{fig:train_infer}
\end{figure*}

In ClearVQA benchmark, GPT-4 \cite{achiam2023gpt} is used to simulate the "\texttt{user}" role, to provide user feedback $\boldsymbol{\xi}_i, i\in \{1,\cdots,n\}$ when VLM poses a clarification question, as shown in Figure~\ref{fig:train_infer} (b). Notably, considering the principle that conversational systems should not expect users to provide complex feedback \cite{deng2023prompting, deng2024towards}, and to facilitate the simulation of user responses for evaluation, we restrict the user feedback $\boldsymbol{\xi}_i$ in (1) to "yes" or "no". 

\subsection{Automatic Data Construction}
\label{data_construction}
In the VQA context, constructing ambiguous visual questions is challenging because it must simultaneously ensure that visual information matters in question answering \cite{goyal2017making}. Some works select examples with significant annotator disagreement from existing VQA datasets as ambiguous visual questions \cite{stengel2023did}. However, the true intent of the questioner in these samples is unknown, making it impossible to simulate user judgment on the correctness of clarification questions posed by VLMs during evaluation. We take a different approach by blurring clearly stated questions with explicit answers from existing VQA datasets to ambiguous ones, with strategies to ensure that such ambiguity aligns with real-world scenes. Some details of the data construction process are provided in Appendix~\ref{appendix:data_details}.

\textbf{Ambiguity Categories.} One issue in automatic data construction is ensuring that the generated data aligns with real-world scenes \cite{guo2024generative}. To address this issue, we focus on some common categories of ambiguity in real-world scenes, rather than allowing the LLM to generate freely. As illustrated in Figure~\ref{fig:ambiguity_category}, ClearVQA includes the following categories of ambiguity: 1) \textbf{Referential ambiguity} \cite{prasad2023rephrase} occurs when the referring expression does not uniquely specify the intended referent; 2) \textbf{Intent underspecification} \cite{bhattacharya2019does} refers to questions with insufficient information to reveal users' requirements, making it hard for VLMs to understand the core and provide a specific answer; 3) \textbf{Spelling ambiguity} \cite{ravichander2021noiseqa} occurs when users misspell some key entities in their questions, potentially changing the meaning, and requiring visual information to disambiguate. 

\textbf{Heuristic preprocessing.} As mentioned previously, we must first collect VQA samples without ambiguity to provide an explicit ground truth intent for evaluation. Previous work \cite{bhattacharya2019does} has shown that discrepancies among annotators can partially reflect question ambiguity, thus we use this metric for filtering. Furthermore, for referential ambiguity and intent underspecification, we aim for the collected samples to be transformable through simple natural language processing (NLP) operations (e.g., removing descriptive clauses), facilitating automatic generation by LLMs. Thus we design some heuristic filtering methods, including question complexity filtering and non-wh-questions filtering. The filtering strategies are detailed in Appendix~\ref{appendix:data_details}.

\textbf{LLM-based ambiguity-clarification question pair generation.} We use in-context learning (ICL) with GPT-4 to generate ambiguous questions and corresponding clarification questions based on the VQA samples collected earlier. Given that current MLLMs exhibit limited multimodal ICL capabilities \cite{monajatipoor2023metavl}, we rely on VLMs to convert images into captions \cite{liu2024seeing}. The construction of spelling error-based ambiguities is achieved by directly prompting GPT-4 with instructions like "\texttt{Replace a named entity with a similar word or typo error.}" Constructing the other two categories of ambiguity is more challenging, as it may not be feasible for every collected VQA sample. For example, images with only one prominent entity do not allow for referential ambiguity. Therefore, for individual samples, we use a crafted prompt containing examples of the two categories, allowing GPT-4 to adaptively choose whether to generate a question with referential ambiguity or intent underspecification. Detailed prompts are provided in Appendix~\ref{appendix:prompt_template}.

\textbf{Human verification and filtering.} To further ensure the quality of the test set, we recruit 12 volunteers to manually select visual questions that exhibit ambiguity and align with real-world scenes, as detailed in Appendix~\ref{appendix:data_quality}. 
The statistical results are shown in Appendix E, demonstrating the data diversity of ClearVQA.


\section{Methodology}
\label{methodology}

We expect the automatically constructed train set in ClearVQA to enable existing VLMs to seek clarification for ambiguous questions and generate answers based on the interactive clarification context. To this end, We implement a two-stage training process, as shown in Figure~\ref{fig:train_infer} (a).

\textbf{Supervised Fine-Tuning (SFT).} As mentioned earlier, the interactive clarification task in the VQA context we claimed involves a visual dialogue \cite{das2017visual} format. We construct SFT data based on the prompt formats of open-source VLMs that support visual dialogue. Meanwhile, since we expect VLMs to answer questions without ambiguity directly, a balanced proportion of unambiguous question-answer pairs is included in SFT data. 

\textbf{Direct Preference Optimization (DPO). } Assuming that based on SFT, VLMs can acquire the ability to perform interactive clarification in the VQA context, we aim to use DPO \cite{rafailov2024direct} to encourage VLMs to seek clarification for ambiguous questions rather than directly answering them, while also avoiding unnecessary clarifications for clearly stated questions. Based on these principles, when constructing DPO preference pairs, the clarification questions annotated previously are preferred over VLMs' direct answers for ambiguous questions. For unambiguous questions, we prompt the SFT-trained VLMs to generate clarification questions as negative samples by using prefixes like "\texttt{Are you referring to}", while the gold answer is used as a positive sample. 

\textbf{Inference.} As shown in Figure~\ref{fig:train_infer} (b), the inference process involves multiple turns of interactive clarification. In each turn, the VLM receives clarification context (including the image and visual question) to either pose a clarification question or generate an answer. "\texttt{User}" (simulated by GPT-4) determines whether a clarification question is posed, and if so, provides user feedback to clarification context. 
Besides, if VLMs fail to capture the true intent of "\texttt{User}" within limited turns \footnote{The maximum turn in interactive clarification is 3. Appendix~\ref{appendix:turns} presents the impact of different turns on interactive clarification performance.}, all clarification context is discarded, and VLMs are prompted to generate an answer directly.

Detailed implementations are shown in Appendix~\ref{appendix:training details}. This implementation is based on LLaVA-1.5 \cite{liu2024improved}.  Thus the VLMs fine-tuned with SFT/DPO could be referred to as Ask-LLaVASFT/DPO.



\section{Experiments} 
\begin{table*}[t]
\small
\centering
\renewcommand{\arraystretch}{1.25} 
\begin{tabular}{lccccccc}
\toprule
\multirow{3}{*}{\textbf{VL Model}} & \multirow{3}{*}{\textbf{Base LLM}} & \multirow{3}{*}{\textbf{Refer}} & \multirow{3}{*}{\textbf{Intent}} & \multirow{3}{*}{\textbf{Spell}} & \multicolumn{2}{c}{\textbf{Overall}} \\
 \cmidrule(lr){6-7}
 &  & & & & \textbf{$\text{VQA}$} & \textbf{$\text{EM}$} \\ 
\midrule
\multicolumn{7}{c}{General Visual Language Models} \\
\midrule
BLIP2 \cite{li2023blip} & Flan-T5-XXL & 19.83 & 18.28 & 23.79 & 20.18 & 25.93 \\
Qwen-VL-7B \cite{bai2023qwen} & Qwen-7B & 20.52 & 18.82 & 24.45 & 20.81 & 28.25 \\
InstructBLIP \cite{dai2023instructblip} & Vicuna-13B & 23.45 & 22.68 & 26.97 & 24.01 & 31.62 \\
LLaVA-1.5-7B \cite{liu2024improved} & Vicuna-7B & 29.83 & 27.91 & 30.74 & 29.28 & 37.88 \\
LLaVA-1.5-13B \cite{liu2024improved}  & Vicuna-13B & 31.74 & 29.40 & 36.34 & 31.93 & 41.21 \\
LLaVA-Next-34B \cite{liu2024LLaVAnext} & Hermes-2-Yi-34B & 33.26 & 30.84 & 35.21 & 32.78 & 42.77 \\
LLaVA-Next-13B \cite{liu2024LLaVAnext} & Vicuna-13B & 33.81 & 31.12 & 38.38 & 33.87 & 43.60 \\
\midrule
\multicolumn{7}{c}{Ambiguity-Resolving Visual Language Models} \\
\midrule
\citet{stengel2023did} & Vicuna-7B & 29.93 & 28.14 & 31.39 & 29.57 & 38.41 \\
\citet{stengel2023did} & Vicuna 13B & 31.81 & 29.52 & 36.68 & 32.09 & 41.28 \\
REPARE \cite{prasad2023rephrase} & Vicuna-13B & 34.04 & 32.21 & 38.45	& 34.40 & 41.87 \\

\midrule
\multicolumn{7}{c}{Our Model (\textit{Tuning on synthetic VQA data with ambiguity})} \\
\midrule
\textbf{$\text{Ask-}\text{LLaVA-7B}$ (SFT)} & Vicuna-7B & 33.57 & 31.06 & 37.23 & 33.47 & 42.34 \\
\textbf{$\text{Ask-}\text{LLaVA-7B}$ (SFT+DPO)} & Vicuna-7B & 34.72 & 32.64 & 39.14 & 34.98 & 44.37 \\
\textbf{$\text{Ask-}\text{LLaVA-13B}$ (SFT)} & Vicuna-13B & 35.25 & 32.75 & 43.28 & 36.23 & 45.19 \\
\textbf{$\text{Ask-}\text{LLaVA-13B}$ (SFT+DPO)} & Vicuna-13B & \textbf{37.06} & \textbf{33.90} & \textbf{45.67} & \textbf{37.92} & \textbf{47.21} \\
\bottomrule
\end{tabular}
\caption{The overall performance of our model, which is based on SFT and DPO using synthesized VQA data with ambiguity, compared with widely used general VLMs and VLMs that have ambiguity-resolving capability. "Refer", "Intent", and "Spell" stands for the VQA Score of VLMs on visual questions with referential ambiguity, intent underspecification, and spelling ambiguity, respectively. "EM" stands for the exact match metric.}
\label{tab:overall_performance}
\end{table*}

\subsection{Experimental Setup}
\label{evaluation_metrics}
To comprehensively evaluate VLMs' ability to handle ambiguity in VQA context through interactive clarification, we employ the following metrics. 

\textbf{VQA Accuracy.} We evaluate the accuracy of VLMs' responses to ambiguous questions based on the widely adopted \textbf{VQA Score} \cite{schwenk2022okvqa} and \textbf{Exact Match (EM)} \cite{rajpurkar2016squad} metrics. This experimental setup allows VLMs to pose clarification questions and generate answers based on user feedback. It is worth noting that user feedback is simulated using GPT-4 by comparing the posed clarification questions with reference, as shown in Appendix~\ref{appendix:prompt_template}. Furthermore, we report the VQA Score of VLMs on visual questions with referential ambiguity, intent underspecification, and spelling ambiguity, respectively. 


\textbf{Ambiguity discrimination accuracy.} We evaluate the ability of VLMs to ask for clarification when handling ambiguous questions, while directly answering clearly stated ones. Therefore, we proportionally mix ambiguous questions (from ClearVQA test set) with clearly stated ones for evaluation, and compute the VLMs' \textbf{precision, recall, and F1 score} \cite{schuff2020f1} based on whether clarification questions are posed.

\textbf{Clarification question quality human assessment.} The quality of clarification questions generated by VLMs is crucial, as it directly affects the reliability from user perspectives \cite{deng2023prompting}. We introduce the following criteria for human evaluation of VLMs generated clarification questions: 1) \textbf{Faithfulness}, which emphasizes that the clarification question must be relevant to the image and not refer to entities not present in the image. 2) \textbf{Reasonableness}, which requires that clarification questions reflect the potential intent of ambiguous questions. 3) \textbf{Clarity}, which stresses that clarification questions must specify a concrete intent or entity, avoiding vague concepts or unclear references. 4) \textbf{Overall quality}. We use the harmonic mean of the three metrics mentioned above to evaluate the overall quality of the clarification questions generated by the VLMs. More human assessment details are shown in Appendix~\ref{appendix:questionnaire}.

\subsection{VQA Accuracy}
\label{experiment:vqa_acc}
\begin{table*}[t]
\small
\centering
\renewcommand{\arraystretch}{1.1} 
\setlength{\tabcolsep}{5pt}
\begin{tabular}{l|cccc|cccc}
\toprule
\textbf{Model}            & \textbf{VQA} & \textbf{VQA}$_\text{D}$ & \textbf{$\Delta$} & \textbf{$\Delta$ (\%)} & \textbf{VQA} & \textbf{VQA}$_\text{Cap}$ & \textbf{$\Delta_\text{Cap}$} & \textbf{$\Delta_\text{Cap}$ (\%)} \\ 
\midrule
\textbf{$\text{Ask-}\text{LLaVA-7B}$ (SFT)} & 33.47  & 30.65  & 2.82 & 9.20 & 33.47  & 31.67  & 1.80 & 5.68 \\
\textbf{$\text{Ask-}\text{LLaVA-7B}$ (SFT+DPO)}  & 34.98  & 30.87  & 4.11 & 13.31 & 34.98  & 32.15  & 2.83 & 8.80 \\
\textbf{$\text{Ask-}\text{LLaVA-13B}$ (SFT)} & 36.23  & 31.59  & 4.64 & 14.69 & 36.23  & 33.12  & 3.11 & 9.39 \\
\textbf{$\text{Ask-}\text{LLaVA-13B}$ (SFT+DPO)}  & 37.92  & 31.61  & 6.31 & 19.96 & 37.92  & 33.33  & 4.59 & 13.77 \\
\bottomrule
\end{tabular}
\caption{Enhancement in VQA accuracy due to interactive clarification. $\Delta$ and $\Delta$ (\%) represent relative and percentage improvement of VQA Score through interactive clarification compared to direct answering (VQA$_\text{D}$). VQA$_\text{Cap}$ refers to VQA Score obtained by replacing interactive clarification contexts of Ask-LLaVA with image captions. $\Delta_\text{Cap}$ and $\Delta_\text{Cap}$ (\%) are relative and percentage improvements in the VQA Score compared to VQA$_\text{Cap}$.}
\label{tab:vqa_enhancement}
\end{table*}

\begin{table}[t]
\centering
\small
\renewcommand{\arraystretch}{1.1} 
\begin{tabular}{l|ccc}
\toprule
\textbf{Model}             & \textbf{Precision} & \textbf{Recall} & \textbf{F1}     \\
\midrule
LLaVA-1.5-13B     & 0.6923    & -      & -      \\
GPT-4V       & 0.8776    & 0.3316 & 0.4813 \\
GPT-4o            & 0.7061    & 0.4447 & 0.5457 \\
\midrule
\multicolumn{4}{c}{Our Method} \\
\midrule

\textbf{$\text{Ask-}\text{LLaVA-}\text{7B}_\text{SFT}$ } & 0.9205    & 0.3691 & 0.5269 \\
\textbf{$\text{Ask-}\text{LLaVA-7B}_\text{DPO}$}  & 0.9362   & 0.4326 & 0.5918 \\
\textbf{$\text{Ask-}\text{LLaVA-13B}_\text{SFT}$ }& 0.9237    & 0.4547 & 0.6094 \\
\textbf{$\text{Ask-}\text{LLaVA-13B}_\text{DPO}$} & \textbf{0.9389}    & \textbf{0.5408} & \textbf{0.6862} \\
\bottomrule
\end{tabular}
\caption{The quantitative results of the VLMs' ability to distinguish whether a question is ambiguous, after performing SFT and DPO on the ClearVQA training set with ambiguous visual questions fully constructed in an automated manner.}
\label{tab:precision_recall_f1}
\end{table}
\textbf{Overall performance.} In Table~\ref{tab:overall_performance}, we present the VQA accuracy of Ask-LLaVA after SFT and DPO, based on the automatically constructed training set from the ClearVQA benchmark.  For comparison, we include mainstream open-source multimodal models such as InstructBLIP and LLaVA-Next \cite{dai2023instructblip, liu2024LLaVAnext}. According to experimental results, Ask-LLaVA surpasses the base model LLaVA by a significant margin in handling ambiguous questions through SFT, benefiting from the interactive clarification capability that open-source VLMs lack. After DPO, the performance of Ask-LLaVA on ambiguous questions improves further, surpassing the base models by 5.70 and 5.99 on the $\text{VQA}$ Score the 7B and 13B scales, respectively. Additionally, Ask-LLaVA outperforms the best-performing open-source VLM, LLaVA-Next-13B, by 4.05 and 3.61 on the $\text{VQA}$ and $\text{EM}$ metrics on visual questions with ambiguity.

Besides, as shown in Table 1, the performance of Ask-LLaVA on visual queries with referential ambiguity, intent underspecification, and spelling ambiguity significantly improves after SFT and DPO. Among these, the VQA Score for spelling ambiguity shows the most substantial improvement, indicating that this ambiguity is more easily clarified by combining text and visual information. In contrast, visual questions with referential ambiguity and intent underspecification often involve multiple possible interpretations of the user's true intent and references. For example, the question in Figure~\ref{fig:clearvqa} could refer to the type of "\texttt{vehicle}" or "\texttt{business}" near the crowd, or it might refer to the "\texttt{pizza}" in the foreground. Therefore, clarifying these two types of ambiguity is more challenging, and the VQA Score improvements obtained through interactive clarification are smaller than visual questions with spelling ambiguity.


We also compare Ask-LLaVA with existing ambiguity-resolving visual–language models. \citet{stengel2023did} train a question-generation model that rewrites ambiguous visual questions. REPARE \cite{prasad2023rephrase} is a purpose-built VQA system that mitigates ambiguity through zero-shot question rephrasing and confidence-based answer selection. Experimental results show that the proposed ClearVQA benchmark remains challenging for current ambiguity-resolving VLMs. With the crafted training data, Ask-LLaVA performs superior to existing ambiguity-resolving VLMs.

\textbf{VQA Accuracy Enhancement.} To confirm that the enhancement in VQA accuracy on ambiguous questions is indeed due to interactive clarification resolving the ambiguity, rather than simply an enhancement in baseline VQA capabilities from additional training, as described in Section~\ref{evaluation_metrics}, we prompt the VLMs to output answers directly without clarification to quantify the gains from interactive clarification. As shown in Table ~\ref{tab:vqa_enhancement}, the interactive clarification capability imparted by SFT and DPO significantly improves VQA accuracy compared to direct answering. For the 7B and 13B models, SFT results in a percentage improvement of 9.32\% and 14.96\% in VQA Score, respectively, while DPO further boosts the gains to 13.31\% and 19.96\%, respectively. 

Furthermore, we replace the interactive clarification context of Ask-LLaVA with image captions in the form of "\texttt{Are you asking about the image that <caption>}", to analyze whether the improvement in VQA accuracy is due to interactive clarification rather than additional context is provided. The user's feedback is fixed as "\texttt{yes}". The VQA Score under this experimental setup, denoted as $\text{VQA}_\text{C}$, and the accuracy improvement $\Delta_\text{Cap}$ and $\Delta_\text{Cap}$ (\%) for Ask-LLaVA relative to $\text{VQA}_\text{C}$ are shown in Table~\ref{tab:vqa_enhancement}. The experimental results indicate that the contribution of a faithful image description to the VQA Score is limited, and the primary performance improvement is still due to the VLM's ability to infer the true intent of visual questions and incorporate user feedback.

\subsection{Discrimination Accuracy of Ambiguity}
\label{experiment:detect_ambiguity}

\begin{table*}[t]
\centering
\small
\renewcommand{\arraystretch}{1.1} 
\begin{tabular}{l|ccccc}
\toprule
\textbf{Model} & \textbf{Recall} & \textbf{Faithfulness} & \textbf{Reasonableness} & \textbf{Clarity} & \textbf{Overall Quality} \\
\midrule
GPT-4V \cite{yang2023dawn}          & 0.3316 & \textbf{1.82} & 1.37 & 0.85 & 1.22 \\
GPT-4o \cite{GPT4o2024}           & 0.4447 & 1.79 & 1.29 & 1.03 & 1.30  \\
\midrule
\multicolumn{6}{c}{Our Method} \\
\midrule
\textbf{$\text{Ask-}\text{LLaVA-}\text{7B}$ (SFT+DPO)}  & 0.4326 & 1.60 & 1.56 & 1.62 & 1.59 \\
\textbf{$\text{Ask-}\text{LLaVA-}\text{13B}$ (SFT+DPO)} & \textbf{0.5408} & 1.65 & \textbf{1.61} & \textbf{1.68} & \textbf{1.65}  \\
\bottomrule
\end{tabular}
\caption{Human evaluation results of the clarification questions generated by VLMs, after SFT and DPO training based on the automatically constructed training set from ClearVQA benchmark, for ambiguous questions, along with a comparison against large-scale close-source VLMs.}
\label{tab:human_evaluation}
\end{table*}

As described in Section~\ref{evaluation_metrics}, we mix the test dataset in ClearVQA benchmark, which contains ambiguous questions, with an equal proportion of randomly sampled non-ambiguous questions to evaluate the ability of VLMs to detect ambiguity. For comparison, in addition to open-source VLMs, we also include large-scale close-source VLMs, GPT-4V \cite{yang2023dawn}, and GPT-4o \cite{GPT4o2024}, which achieve state-of-the-art performance across multiple cross-modal tasks. We use prompts such as "\texttt{If feel ambiguous, please ask ... to clarify. Otherwise, you can answer directly}." to guide the VLMs in identifying ambiguity in the visual questions. However, as shown in Table~\ref{tab:precision_recall_f1}, even with crafted prompts, open-source models like LLaVA struggle to seek clarification for ambiguous questions. This limitation arises because, unlike close-source VLMs, they have not been pre-trained and preference-optimized on data with sufficiently diverse formats.

Based on the automatically constructed training set within ClearVQA benchmark, Table~\ref{tab:precision_recall_f1} quantifies Ask-LLaVA 's ability to identify whether a question contains ambiguity. The experimental results show that even with simple SFT, owing to our well-designed data construction method, Ask-LLaVA is capable of recognizing ambiguity in visual questions, surpassing the performance of GPT-4V and even GPT-4o in this task. After DPO, this ability is further enhanced, with $\text{Ask-}\text{LLaVA-13B}_\text{DPO}$ achieving an F1 score of 0.6862, surpassing GPT-4o by 0.14.

\subsection{Quality of Clarification Questions}
In Section~\ref{experiment:detect_ambiguity}, the accuracy and recall of Ask-LLaVA in identifying ambiguous questions are analyzed. However, even if ambiguity is correctly identified, failing to generate appropriate clarification questions would still render the model unreliable from the user’s perspective. Therefore, we assess this reasonableness based on the human evaluation criteria proposed in Section~\ref{evaluation_metrics} and detailed in Appendix~\ref{appendix:questionnaire}. For comparison, we employ large-scale closed-source VLMs, GPT-4V and GPT-4o. We prompt GPT-4V and GPT-4o to generate clarification questions for ambiguous visual questions using prompts like "\texttt{The question is difficult to answer due to its ambiguity. Please generate a clarification question ..., like ...}". We randomly sample 100 clarification questions generated by Ask-LLaVA, GPT-4V, and GPT-4o, and recruit volunteers to perform human assessment. The results are shown in Table~\ref{tab:human_evaluation}.

The experimental results indicate that although Ask-LLaVA performs slightly below the well-trained GPT-4V and GPT-4o in terms of faithfulness to the image, it reaches a comparable level. More importantly, Ask-LLaVA significantly outperforms GPT-4V and GPT-4o in the metrics of reasonableness and clarity. This is because Ask-LLaVA generates outputs like "\texttt{Are you asking in which major city the woman is most likely sitting outdoors?}" in Figure~\ref{fig:ambiguity_category}, which provides a clear inference of the question's potential intent. In contrast, since GPT-4V and GPT-4o have not been specifically trained, their outputs often include vague statements like "\texttt{If there is any specific context or part of the image you are referring to, please let me know ...}", which require users to provide lengthy clarifications themselves. Moreover, based on the reasonableness metric that comprehensively considers faithfulness, reasonableness, and clarity, we conclude that the training data, generated through our crafted automated data construction method,  enables VLMs to maintain robustness in handling ambiguous visual questions from a human perspective. This capability even surpasses that of large-scale models like GPT-4V and GPT-4o.

\subsection{Performance on Real-World Sences}
\begin{table*}[t]
\centering
\small
\renewcommand{\arraystretch}{1.1} 
\setlength{\tabcolsep}{5pt}
\begin{tabular}{l|cccccc}
\toprule
\textbf{Model} & \textbf{VQA} & \textbf{VQA}$_\text{D}$ & \textbf{$\Delta$} & \textbf{$\Delta$ (\%)} & \textbf{Recall} & \textbf{Overall Quality} \\
\midrule
$\text{Ask-}\text{LLaVA-7B}_\text{DPO}$ & 36.91 & 33.74 & 3.17 & 9.39 & 0.3225 & 1.53 \\
$\text{Ask-}\text{LLaVA-13B}_\text{DPO}$ & 39.67 & 35.25 & 4.42 & 12.51 & 0.4125 & 1.56 \\
\bottomrule
\end{tabular}
\caption{The performance of the VLM, after SFT and DPO, on ambiguous questions in real-world scenes derived from VQAv2. $\Delta$ and $\Delta$ (\%) represent relative and percentage improvement of VQA Score through interactive clarification compared to direct answering (VQA$_\text{D}$). Overall quality: Clarification quality human assessment metric.}
\label{tab:real_scene}
\end{table*}

We further evaluate the performance of Ask-LLaVA on real-world ambiguous questions posed by humans, rather than synthetic ambiguous questions. Specifically, the evaluation questions are derived from VQAv2 \cite{goyal2017making}, a widely used dataset comprising diverse human-posed questions. Following the manual filtering approach in \cite{ni2024visual}, we sample 400 ambiguous questions from VQAv2 test set. As described in Section~\ref{evaluation_metrics}, GPT-4 simulates user feedback by comparing the clarification questions posed by VLMs with reference. For collected ambiguous questions in real-world scenes, we recruit volunteers to annotate clarification question references based on ground truth answers manually.

The experimental results in Table~\ref{tab:real_scene} show that even in real-world scenes, our method can identify ambiguity, generate reasonable clarifications, and improve the final VQA accuracy, with examples provided in Appendix~\ref{appendix:Case Study}. This indicates that the data synthesized by our crafted automatic pipeline aligns well with real-world scenes. However, due to inevitable factors like ClearVQA benchmark cannot include all ambiguity categories, Ask-LLaVA’s recall for ambiguity and the improvement in VQA Score on real-world scenes are relatively lower compared to those on the ClearVQA test set.

\section{Related Works}
\textbf{Visual Question Answering (VQA).}  The VQA task requires VLMs to generate a single answer to visual questions based on images. Early research focused on integrating multimodal features for answer generation \cite{wang2017explicit, anderson2018bottom}. Later, VLMs adopted a pre-training and fine-tuning approach \cite{liang2024document, jian2024large, jing2024dq, zhang2025understand, guo2025crop, jian2025lookagainthinkslowly}, achieving promising results on various VQA benchmarks \cite{alayrac2022flamingo, li2023blip}. More recently, large VLMs or LLMs have shown advanced performance using prompt engineering like in-context learning, without extensive training \cite{shao2023prompting, sun2024generative}. As the VQA field gradually evolves, datasets focusing on various real-world scenes are proposed, VQAv2 \cite{goyal2017making} is the largest VQA dataset, including diverse VQA samples. 
OK-VQA \cite{marino2019ok} and A-OKVQA \cite{schwenk2022okvqa} emphasize VLMs' ability to understand fine-grained knowledge beyond commonsense. A-OKVQA further focuses on knowledge-based reasoning.
These efforts provide a broad and valuable resource for our research.  

\textbf{Question Ambiguity.} In NLP filed \cite{zhang2024navigating, zhang2023comprehensive}, ambiguity in instructions or questions is extensively explored \cite{schutze1995ambiguity, min2020ambigqa, futeral2023tackling, ye2025cpoaddressingrewardambiguity}. In multimodal and VQA context, \citet{bhattacharya2019does} claim that the underspecification of questions in VQA is a major contribution to discrepancies among annotators. 
\cite{pezzelle2023dealing} highlights that ambiguity is a critical source of errors in visual language tasks. 
\cite{stengel2023did} collect ambiguous questions from existing VQA datasets and propose a rephrased model to mitigate ambiguity. \citet{prasad2023rephrase} focuses on leveraging existing LLMs for VLMs, facilitating rephrasing in a zero-shot approach, making it applicable to real-world scenes, and achieving performance enhancements in several VQA datasets. \citet{ni2024visual} constructs a multimodal, multi-turn reasoning framework that mimics human cognitive processes to handle ambiguity.

\textbf{Selective Question Answering.} Recent studies advocate for models to refuse questions they do not know, thereby enhancing their reliability \cite{srinivasan2024selective, cheng2024can}. In VQA context, \citet{whitehead2022reliable} employs a selection function, to make the first exploration of deep models with a rejection option. \citet{khan2024consistency} employs a proxy model to rephrase visual questions and refuse answers when different rephrasings yield inconsistent outputs, implementing selective prediction in a black-box manner. These works inspire our investigation into how VLMs can selectively ask for clarification. However, our focus is more challenging, as it involves not only rejecting to answer but also emphasizing the reasonableness of clarification questions posed to users.

\section{Conclusion}
We propose ClearVQA benchmark, for evaluating VLMs' ability to handle ambiguity in visual questions through interactive clarification. ClearVQA benchmark focuses on three common ambiguity categories in visual questions and involves several VQA scenarios. Furthermore, ClearVQA includes a training set constructed based on a crafted and automated pipeline. 
Experimental results show that VLMs trained on this synthetic data can generate reasonable clarification questions for ambiguous questions, thereby generating more accurate and specific answers based on user feedback.

\section*{Limitations}
Firstly, as the first attempt to address visual question ambiguity through an interactive approach, we focus on user responses limited to "yes" or "no" for simplicity, enabling simulating user feedback with intelligent agents like GPT-4. This may not fully capture the complexity of interactive clarification in real-world scenes. We plan to improve the evaluation process to support more diverse user feedback in future work. Secondly, although our approach grants VLMs interactive clarification ability from scratch, they seldom generate truly new clarification questions after misreading user intent. Future work might address this with reflection \cite{cheng2024vision}, Monte Carlo tree search \cite{chen2024alphamath}, and other advanced reasoning techniques \cite{sun2025ktae, chen2025lr, ren2025towards}. Thirdly, due to computational constraints, we limited our exploration of VLMs’ interactive clarification capabilities to the 7B and 13B scales.

\section*{Acknowledgments}
We thank our colleagues Junhong Wu and Jianghao Chen for their helpful suggestions during the writing stage of the paper. We also gratefully acknowledge the volunteers who contributed to the data construction process. Finally, we thank all reviewers for their detailed reviews and insightful comments. 

This work is supported by National Key R\&D Program of China 2022ZD0160602 and the Strategic Priority Research Program of Chinese Academy of Sciences under Grant XDA04080400.

\bibliography{acl_latex}

\clearpage

\appendix
\section{Prompt Template}
\label{appendix:prompt_template}
In this section, we provide detailed prompt templates for constructing questions with referential ambiguity, intent underspecification, and spelling ambiguity as follows.

\begin{tcolorbox}[breakable, title=Prompt for Constructing Referential Ambiguity and Intent Underspecification]
    \small \texttt{I'll give you a question for an image, the corresponding answer, and a textual description of the image, please complete the following task for me: Blur the question so there is ambiguity in the question. You will also need to pose a clarification question that explains the ambiguity of the previously blurred question. You have the following options: 1. blur the intent of the problem, e.g., a) and b) below; 2. replace the entities that appear in the problem with an ambiguous reference, e.g., c) and d)  below.    \\}

    \textcolor{blue}{\small \texttt{Examples: \\a) Question: Why would we suspect that these bears are male and female? \\ Caption: a couple of bears sitting on top of a rock\\ 
    Answer: kiss \\ Blurred question: Why would we suspect these bears are different? \\
    Clarification question: Do you mean the reason we suspect these bears are male and female?  \\
    Ambiguous Type: 1  
    \\}}
    
    \textcolor{blue}{\small \texttt{b) Question: This type of bus can be found in what popular city? \\
    Caption: A trolley car traveling down a city street\\ 
    Answer: kiss \\ Blurred question: Where this type of bus can be found? \\
    Clarification question: Do you want to know this type of bus can be found in what popular city? \\
    Ambiguous Type: 1  
    \\}}
    
    \textcolor{blue}{\small \texttt{c) Question: Name the material used to make these umbrella shown in this picture? \\ Caption: A group of people walking through a park with umbrellas hanging from the trees\\ 
    Answer: paper. \\ Blurred question: What material used to make them? \\
    Clarification question: Are you referring to these colorful umbrellas in the picture? \\
    Ambiguous Type: 2 \\}}

    \textcolor{blue}{\small \texttt{c) Question: What auction company is accessible only via the item featured in this photo? \\ Caption: a person typing on a laptop computer at a desk.\\ 
    Answer: ebay  \\ Blurred question: What auction company is }}
    \textcolor{blue}{\small \texttt{ accessible only via this method? \\
    Clarification question: Are you referring to the auction company accessible only via laptop computer?\\
    Ambiguous Type: 2\\}}

    \small \texttt{Question: <Question> \\}
    \small \texttt{Caption: <Caption> \\}
    \small \texttt{Answer: <Answer>}
\end{tcolorbox}

\begin{tcolorbox}[breakable, title=Prompt for Constructing VQA Samples with Spelling Ambiguity]
    \small \texttt{I'll give you a question for an image, please complete the following task for me: Replace a named entity in the question with a similar word. You also need to ask a question to clarify the misleading effect of the above modification. \\}
    
    \textcolor{blue}{\small \texttt{Examples: \\Question: Why would we suspect that these bears are male and female? \\ Blurred question: Why would we suspect that these beers are male and female?
    \\ Clarification question: For the "beers" in the picture, do you mean "bears"? \\}}
    
    \textcolor{blue}{\small \texttt{Question: This type of bus can be found in what popular city? \\ 
    Blurred question:  This type of bush can be found in what popular city? \\
    Clarification question: By "bush" in your question, do you mean "bus" in the picture? \\}}

    \small \texttt{Question: <Question>}
\end{tcolorbox}

Additionally, we offer prompts used to evaluate the ability of VLMs to perform interactive clarification in VQA scenarios, where the LLM is instructed to assess whether the clarification question proposed by the VLM aligns with the user’s original intent and to simulate user feedback, based on reference clarification question.

\begin{tcolorbox}[breakable, title=Prompt for Constructing Simulate User Feedback]
    \small \texttt{Are these two sentences semantically similar (yes / no). \\}

    \textcolor{blue}{\small \texttt{Input: \\Sentence 1: Are you asking about the number on the front of the bus? \\ Sentence 2: Are you asking for the 4-digit number visible on both the front of the bus?\\ 
    output: \\ yes\\}}
    
    \textcolor{blue}{\small \texttt{Input: \\Sentence 1: Are you asking which cow has the most leaves on its back? \\ Sentence 2: Are you asking about the tree that has the most branches with leaves?\\ 
    output: \\ no\\}}

    \textcolor{blue}{\small \texttt{Input: \\Sentence 1: For the "houses" in the picture, do you mean "horses"?\\ Sentence 2: When you refer to "houses" in the question, are you instead referring to "horses" seen in the picture?\\ 
    output: \\ yes\\}}

    \textcolor{blue}{\small \texttt{Input: \\Sentence 1: When you mention "planting" in your question, are you referring to the "plant" on the counter?\\ Sentence 2: For the "planting" in your question, do you mean "painting" that is on the counter?\\ 
    output: \\ no \\}}

    \small \texttt{Input: \\}
    \small \texttt{Sentence 1: <Clarification Question> \\}
    \small \texttt{Sentence 2: <Reference> \\}
    \small \texttt{Output:}
\end{tcolorbox}

Finally, we include prompts used to evaluate the impact of interactive clarification on VQA accuracy, designed to guide VLMs to directly answer questions without seeking clarification.

\begin{tcolorbox}[breakable, title=Prompt for Guide VLMs to Answer Directly ]
    \small \texttt{USER: <image>$\backslash$nGiven the image, answer the following question with no more than three words. <Question> ASSISTANT:}
\end{tcolorbox}

\section{Questionnaire for Clarification Human Evaluation}
\label{appendix:questionnaire}
We provide the questionnaire used for human evaluation of the reasonableness of clarification questions generated by VLMs as follows.

\begin{tcolorbox}[breakable, title=Questionnaire for Clarification Question Human Assessment]
    \small \texttt{This is a clarification question posed by the visual language model in response to the ambiguous visual question "<Question>": <Clarification>. Please assess the reasonableness of the clarification question based on the following criteria. (For each metric below, please select a score from {1, 2, 3}, where 1 means poor, 2 means acceptable, and 3 means good.)\\ \\
    This is the image corresponding to the Visual question: <image>
    \\}
    
    \small \texttt{\\1) Faithfulness: This metric evaluates whether the clarification question is relevant to the image. If the question involves entities not present in the image or refers to unrelated concepts, this score should be lower. Otherwise, assign a higher score. \\}
    \\
    \small \texttt{2) Reasonableness: This metric measures whether the clarification question reflects the potential intent of the original ambiguous question. Below are two positive examples with a score of 2 and one negative example with a score of 1. \\}
    \begin{center}
        \includegraphics[width=0.75 \textwidth]{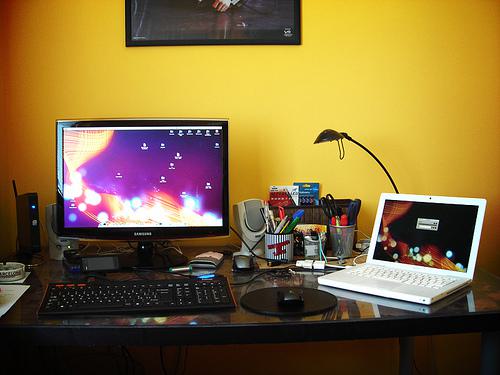} 
    \end{center}

    \textcolor{blue}{
    \small \textit{Question with ambiguity: What are these little things called?  \\
    Positive example 1: Are you asking about the little things on the screen of the device on the left? \\}}
    
    \textcolor{blue}{\small \textit{Positive example 2: Are you referring to the stationery in those cylindrical containers? \\
    Negative example: Are you referring to the two computers in the image?\\} }
    \\
    \small \texttt{\\3) Clarity: The clarification question must specify a concrete intent or entity, avoiding vague concepts or unclear references, such as "If there's another specific detail you're referring to, please let me know!" or "Could you please clarify the details or confirm the object of interest in the image?" } 
    
\end{tcolorbox}

\section{Data Quality Human Verification}
\label{appendix:data_quality}
\begin{table*}[ht]
\centering
\small
\renewcommand{\arraystretch}{1.4} 
\begin{tabular}{l|ccc|c}
\toprule
\textbf{Ambiguity Category} & \textbf{Ambiguity} & \textbf{Clarification Usefulness} & \textbf{Reality} & \textbf{Overall Acceptable} \\
\midrule
Referential Ambiguity       & 96.1\% & 94.9\% & 83.4\% & 82.1\% \\
Intent Underspecification    & 92.8\% & 89.4\% & 81.3\% & 80.9\% \\
Spelling Ambiguity          & 95.6\% & 94.3\% & 76.4\% & 75.8\% \\
\bottomrule
\end{tabular}
\caption{
Human evaluation and filtering of ClearVQA were conducted from three aspects: Whether ambiguity is introduced (ambiguity), whether the clarification question is helpful (clarification usefulness), and whether the data aligns with real-world scenes (reality). "Overall acceptable" refers to the proportion of data meeting all quality filtering criteria.}
\label{tab:human_verfication}
\end{table*}

We employed several human verification strategies to ensure the quality of data automatically constructed using models like GPT-4. The human data quality validation focuses on three key aspects.
\begin{itemize}
  \item \textbf{Ambiguity}: The constructed ambiguous question should introduce actual ambiguity, making it harder to answer than the original clearly stated question.
  \item \textbf{Clarification usefulness}: The reference clarification question should be useful to eliminate the ambiguity of the constructed question.
  \item \textbf{Reality}. The constructed question must align with real-world scenes, i.e., those that might be raised by the user.
\end{itemize}
Human data quality verification was performed at two stages of the proposed automated data construction process.

\textbf{Human evaluation for prompt refinement.} During the data construction prompt design stage, we first sample ambiguous-clarification question pairs, and manually assess whether their quality meets the requirements for \textbf{ambiguity} and \textbf{clarification usefulness}. If the requirements are not met, the prompt is adjusted. 

\textbf{Human verification of ClearVQA test set.} To further ensure the quality of the test set, we recruited 12 volunteers to manually select visual questions of acceptable quality. Unlike the prompt refinement stage, this human quality validation stage not only focuses on \textbf{ambiguity} and \textbf{clarification usefulness} but also considers \textbf{reality}. This means that the ambiguous questions must align with real-world scenes. We designed the following questionnaire for human filtering of the test set. The three questions in the questionnaire correspond to the three aspects of data quality we focus on. The evaluation results are shown in Table~\ref{tab:human_verfication}. For a sample to be considered acceptable, the annotator must answer "yes" to the quality of all three questions in the questionnaire.

\begin{tcolorbox}[breakable, title=Questionnaire for Human Quality Filtering of Test Set in ClearVQA]
    \small \texttt{Image: <image>\\}
    \small \texttt{Ground Truth Answer: <GTA> \\}
    \small \texttt{Original Question: <OQ> \\}
    \small \texttt{Constructed Ambiguous Question: <CAQ> \\}
    \small \texttt{Reference Clarification Question: <RCQ> \\}

    \small \texttt{For the given image, original question, ground truth answer, constructed ambiguous question, and reference clarification question, please respond to the following:\\ } 

    \small \texttt{1) Ambiguity. Does the constructed ambiguous question introduce actual ambiguity compared to the original question, making it impossible to provide an answer or suggest other reasonable answers beyond the ground truth answer? (Yes / No)\\}

    \small \texttt{2) Does the reference clarification question provide additional information that helps generate an answer to the constructed ambiguous question? (Yes / No)\\}

    \small \texttt{3) In real-world scenes, is it possible that the user could express the intent of the original question through the ambiguous question? (Yes / No)\\}
\end{tcolorbox}

The human evaluation results in Table 6 show that our crafted pipeline can automatically generate questions with actual ambiguity and useful clarification questions. Furthermore, most of these ambiguous questions align with real-world scenes, i.e., these questions could be raised by users.

\section{Training Details}
\label{appendix:training details}
We selected the checkpoints \texttt{LLaVA-1.5-7b-hf} and \texttt{LLaVA-1.5-13b-hf} from Huggingface as the base models during training. Table~\ref{tab: training details sft} presents the training parameters for the SFT phase, where the learning rate is set to $2 \times 10^{-4}$, and specifically, the learning rate for the multimodal projector is $2 \times 10^{-5}$. We apply LoRA \cite{hu2021lora} for SFT and DPO. The LoRA training hyperparameters are $\text{Lora}_r =128$ and $\text{Lora}_\alpha =256$. The DPO training parameters, applied after the SFT stage, are shown in Table~\ref{tab: training details dpo}, with a learning rate of $5 \times 10^{-7}$, including the multimodal projector. In the SFT and DPO stages, we used 1 and 2 Nividia A100 GPU for training, respectively.

For VLMs with capabilities to pose clarification questions after SFT, we sample the incorrect clarification questions they generate that do not align with the reference (judged by GPT-4). Therefore, we can construct multi-turn dialogue data in the format of "\texttt{Error clarification; 'No' feedback; Reference clarification; 'Yes' feedback; Answer}" Data with such format enables VLMs to generate different clarification questions when the "\texttt{user}" responds with "\texttt{No}", thereby equipping VLMs with the ability for multi-turn interactive clarification.

It is worth noting that the prompt used to elicit direct answers from VLMs in Appendix~\ref{appendix:prompt_template} is frequently employed in the training and VQA capability assessment of VLMs like LLaVA \cite{liu2024improved}. We take additional training setup to further avoid any decline in VQA performance due to the use of this prompt after training with ClearVQA data, which could lead to unfair assessments of the accuracy improvements due to interactive clarification. specifically, we phrase VQA samples from datasets like VQAv2 with this prompt and incorporate them into the training data.

\begin{table}[t]
\small
\centering
\renewcommand{\arraystretch}{1.2}
\begin{tabular}{c|c}
\toprule
Hyper-parameters & Value \\ \hline
Training steps & 300 \\
Batch size & 64 \\
Warmup ratio & 0.03 \\
Gradient accumulation & 2 \\
Learning rate scheduler & Cosine \\
Learning rate & $2\times10^{-4}$ \\
Projector learning rate & $2\times10^{-5}$ \\
$\text{Lora}_r$ & 128 \\
$\text{Lora}_\alpha$ & 256 \\
GPUs & 2 \\
Optimizer & Adam \\ \bottomrule
\end{tabular}
\caption{The hyper-parameters used during SFT of VLMs using the training set in ClearVQA benchmark.}
\label{tab: training details sft}
\end{table}

\begin{table}[t]
\small
\centering
\renewcommand{\arraystretch}{1.2}
\begin{tabular}{c|c}
\toprule
Hyper-parameters & Value \\ \hline
Training steps & 200 \\
Batch size & 32 \\
Warmup ratio & 0.03 \\
Gradient accumulation & 2 \\
Learning rate scheduler & Cosine \\
Learning rate & $5\times10^{-6}$ \\
$\text{Lora}_r$ & 128 \\
$\text{Lora}_\alpha$ & 256 \\
GPUs & 1 \\
Optimizer & Adam \\ \bottomrule
\end{tabular}
\caption{The hyper-parameters used during DPO of VLMs using the training set in ClearVQA benchmark.}
\label{tab: training details dpo}
\end{table}

\section{Data Details}
\label{appendix:data_details}

\begin{figure*}[t]
  \centering
  \includegraphics[width=0.99\textwidth]{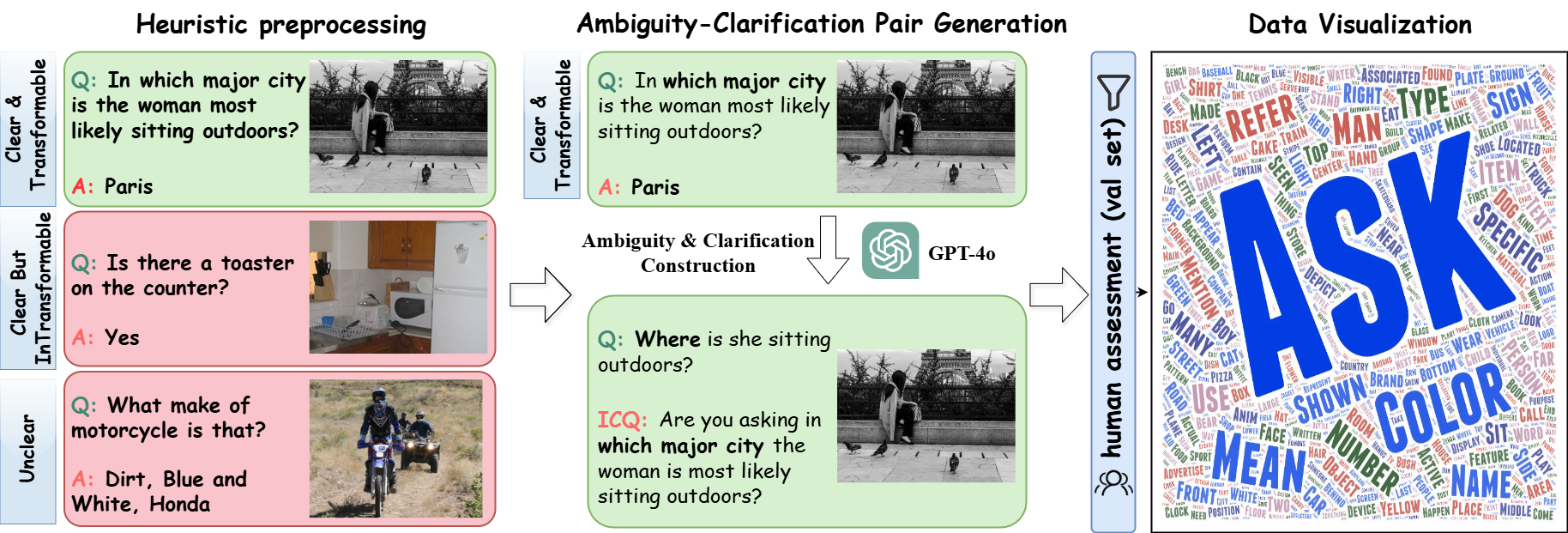}
  \caption{The automated construction process of ambiguity-clarification question pairs. Here, ICQ refers to the ideal clarification question generated by GPT-4, used for training and evaluation.}
  \label{fig:data_construction}
\end{figure*}

We present an intuitive flowchart in Figure~\ref{fig:data_construction} for the automated data construction process introduced in Section~\ref{data_construction}. Using heuristic preprocessing, we filter clearly stated VQA samples that are easily transformable into ambiguous ones. For these VQA samples, we prompt GPT-4 to introduce ambiguity by removing or blurring key information through simple NLP operations. Meanwhile, a reference clarification question is generated based on the clearly stated original question. More specific details are provided below.

\begin{figure}
  \centering
  \includegraphics[width=0.49\textwidth]{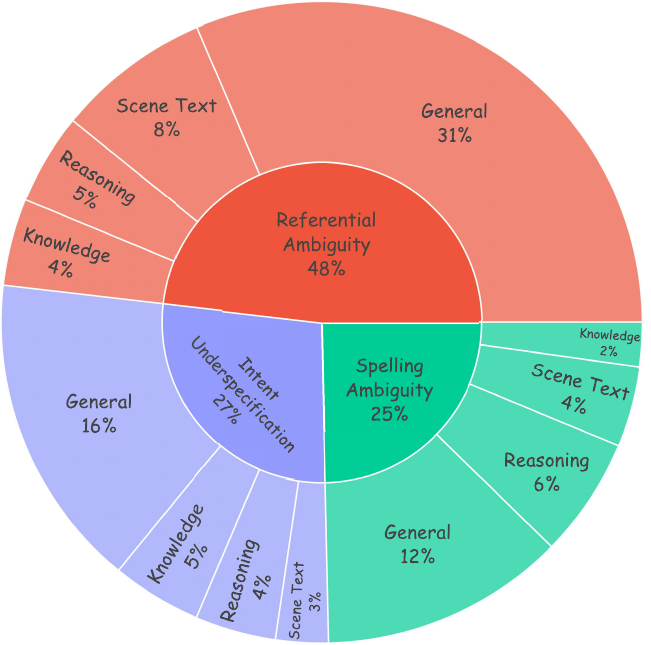}
  \caption{The data diversity in ClearVQA, illustrated by data distribution in the ClearVQA test set. ClearVQA focuses on three common categories of ambiguity. Each ambiguity category includes a variety of VQA scenarios, including general, knowledge, reasoning, and scene text.}
  \label{fig:pie}
\end{figure}

\begin{table*}[t]
\small
\centering
\renewcommand{\arraystretch}{1.4}
\begin{tabular}{l|cccc|cc}
\toprule
\textbf{Ambiguity Category}    & \textbf{General} & \textbf{Knowledge} & \textbf{Reasoning} & \textbf{Scene Text} & \textbf{Train} & \textbf{Test} \\ \midrule
\textbf{Referential Ambiguity} & 4580             & 824            & 2188            &    1651               & 7305          & 1938        \\
\textbf{Intent Underspecification} & 3316          & 661                 & 1276             &     635          & 4793           & 1095       \\
\textbf{Spelling Ambiguity}    & 3120             & 531                 & 1567               & 1011                & 5236           & 992        \\ 
\textbf{Overall}               & 11016            & 2016                & 5031               & 3297                & \textbf{17334}          & \textbf{4025}         \\ \bottomrule
\end{tabular}
\caption{ClearVQA Benchmark Data Statistics. The various VQA scenarios, like knowledge and scene text, are derived from different existing VQA datasets, as detailed in Appendix~\ref{appendix:data_details}.}
\label{tab:ambiguity}
\end{table*}

\begin{figure*}[t]
  \centering
  \includegraphics[width=0.99\textwidth]{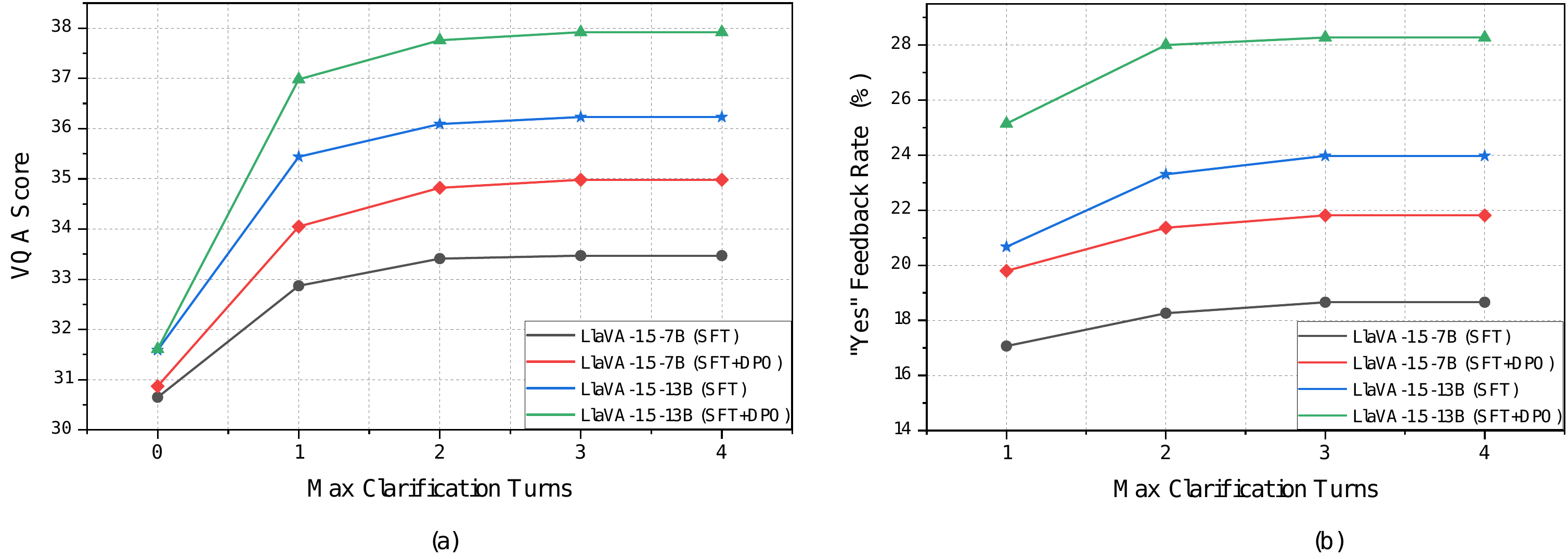}
  \caption{The impact of clarification turns on the performance of interactive clarification. (a) VQA Score after different numbers of clarification turns. Turn 0 represents the VQA Score when VLMs are prompted to answer directly without clarification. (b) The proportion of test set samples whose ambiguity is successfully clarified ("\texttt{user}" feedback is "\texttt{Yes}") after different clarification turns.}
  \label{fig:turns}
\end{figure*}

\textbf{Detailed heuristic preprocessing.} As described in Section~\ref{data_construction}, we first need to collect VQA samples without ambiguity to provide a clear, well-defined intent during evaluation. We select samples based on the \textbf{degree of annotator disagreement}. In many VQA datasets, the ground truth is derived from an answer list annotated by 10 individuals. If none of the answers appears more than three times, we exclude the sample. Additionally, for the two categories of ambiguity other than spelling errors, we aim for the collected samples to be achievable through simple NLP operations, making it easier for the LLM to annotate automatically. To achieve this, we designed several heuristic filtering methods: 1) \textbf{Length filtering}. Questions with more words typically provide more detail to express user intent clearly, and simply removing some key information can make the question ambiguous. Therefore, we apply a word count threshold to filter for the desired questions mentioned earlier. In this work, we exclude VQA samples with fewer than 13 words. 2) \textbf{Non-special interrogative sentence filtering}. Generally, open-ended questions like special interrogative sentences are more likely to involve ambiguity, while yes/no or multiple-choice questions have limited answering space and are less prone to ambiguity. We apply regularization strategies, such as filtering questions where the answers are "\texttt{yes}," "\texttt{none}," "\texttt{no}," to exclude yes/no or multiple-choice VQA samples, retaining only those that are special interrogative sentence. When designing the filtering strategy, we sample the filtered data to check if it meets the expected criteria. If not, we adjust the strategy accordingly.

\textbf{Data without ambiguity in ClearVQA.} In section~\ref{evaluation_metrics}, we use samples without ambiguity for evaluating VQA accuracy in abstain setup and ambiguity discriminative ability. It is worth noting that these non-ambiguous samples were derived from the data filtered based on discrepancies among annotators, as discussed in Section ~\ref{data_construction}. We did not directly use the original questions that were ultimately employed to construct the ambiguous questions, because the heuristic preprocess described before tends to filter questions with fewer words. This approach helps prevent VLMs from performing well on certain metrics merely by identifying longer questions as unambiguous.

\textbf{Diversity and data statistics.} We collect original question samples from various existing VQA datasets, covering a wide range of VQA scenarios, to ensure the diversity of the ClearVQA benchmark. Specifically, the samples for the general VQA scenario in Table~\ref{tab:ambiguity} are derived from the widely-used VQAv2 \cite{goyal2017making}, a dataset encompassing visual understanding, counting, and entity classification questions. To incorporate fine-grained knowledge beyond common sense, we sample from OK-VQA \cite{marino2019ok} for the knowledge VQA scenario. Additionally, we source reasoning VQA scenario samples from A-OKVQA \cite{schwenk2022okvqa}, a dataset emphasizing reasoning based on commonsense or fine-grained knowledge in the image. Besides, scene text VQA scenario samples are collected from ST-VQA \cite{biten2019scene}. The data distribution of ClearVQA across different ambiguity categories and VQA scenarios is shown in Table~\ref{tab:ambiguity} and Figure~\ref{fig:pie}. This statistical result demonstrates the significant data diversity of ClearVQA benchmark.

\section{The Impact of clarification turns.}
\label{appendix:turns}

We designed experiments to analyze how increasing clarification turns affects interactive clarification performance. Specifically, we recorded the VQA accuracy and the proportion of samples where the user's intent was correctly inferred (feedback as "\texttt{Yes}") after different numbers of turns, as shown in Figure~\ref{fig:turns}. The results indicate that the most significant improvements in VQA accuracy and the proportion of correctly inferred user intent occur in the first turn, with smaller gains from the second and third turns. This result demonstrates that accurately inferring user intent enhances VQA accuracy. However, no further performance improvement is observed when the number of clarification turns exceeds three. This suggests that, despite learning to handle ambiguity through interactive clarification, LLaVA's reasoning ability remains limited.

\section{Case Study}
\label{appendix:Case Study}

\begin{figure*}[t]
  \centering
  \includegraphics[width=0.99\textwidth]{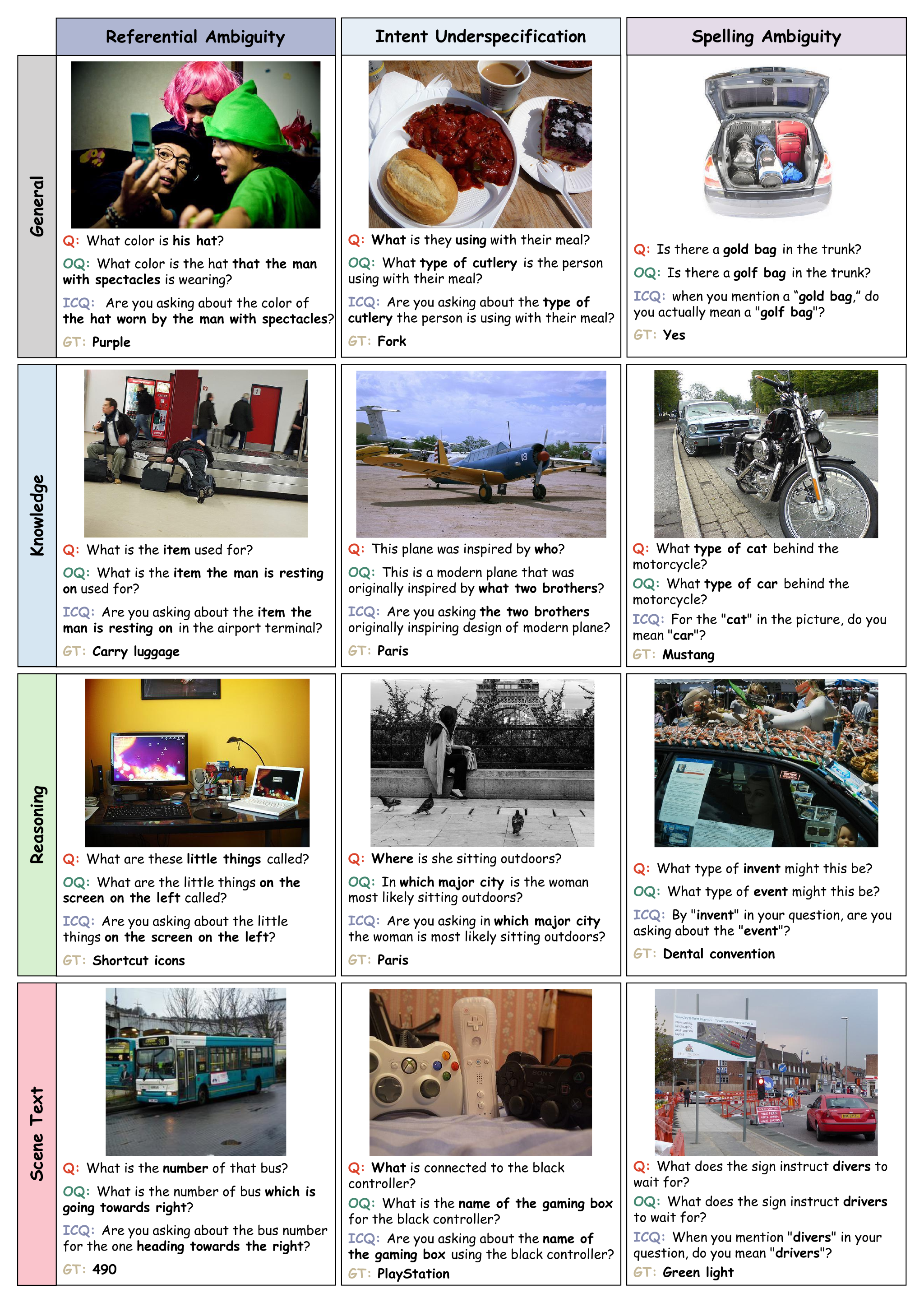}
  \caption{Examples of different VQA scenarios and different ambiguity categories in ClearVQA benchmark. Q: question. IQ stands for the user’s intended question. ICQ represents the ideal clarification question. GT stands for the ground truth answer.}
  \label{fig:clearvqa_case}
\end{figure*}

\begin{figure*}[t]
  \centering
  \includegraphics[width=0.98\textwidth]{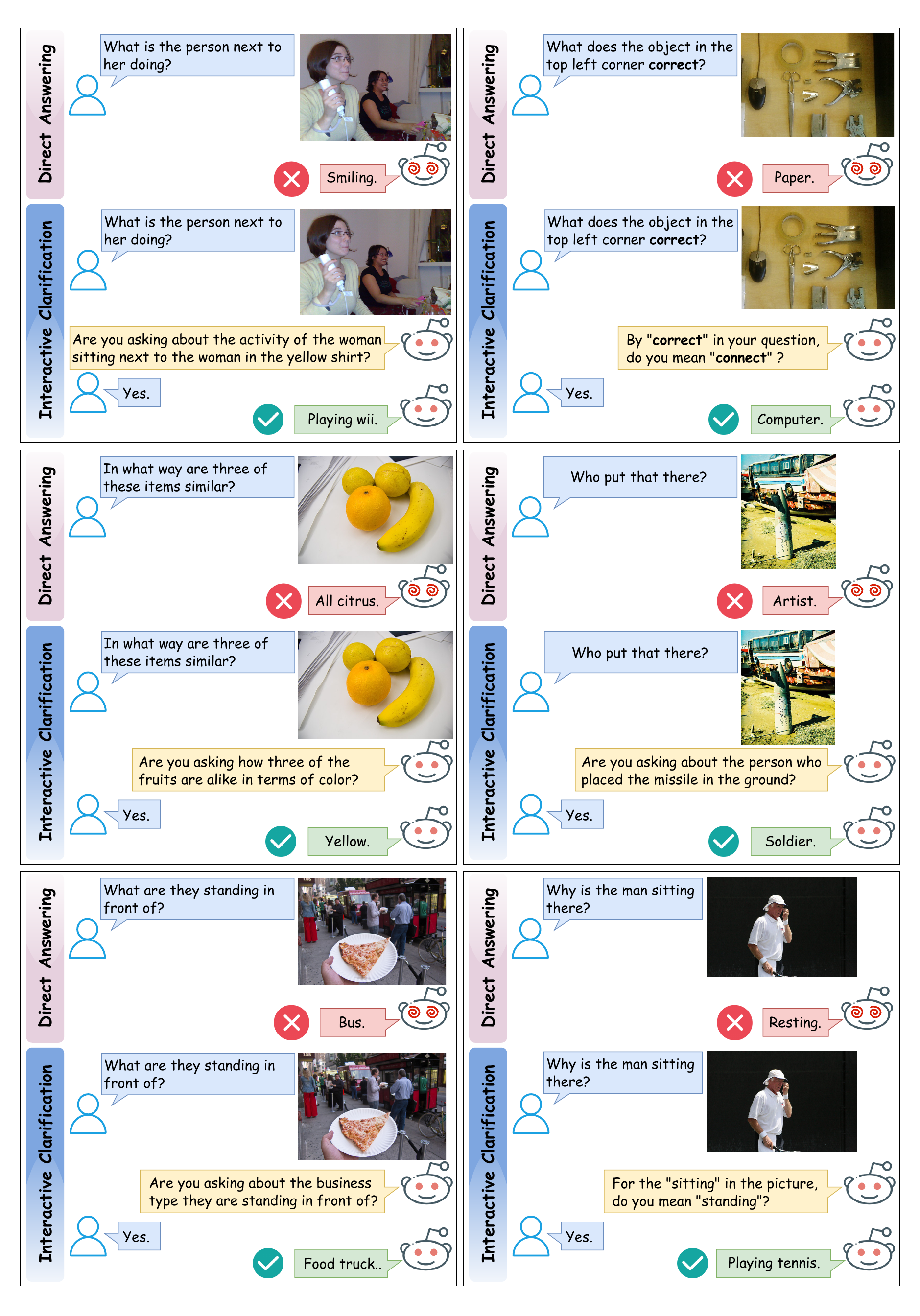}
  \caption{Examples demonstrating the performance of our emphasized interactive clarification in real-world scenes. These visual questions are posed by humans and derived from the VQAv2 dataset.}
  \label{fig:case_study}
\end{figure*}

In Figure~\ref{fig:clearvqa_case}, we present cases from the ClearVQA benchmark across various VQA scenarios and ambiguity categories. Furthermore, examples in Figure~\ref{fig:case_study} illustrate how VLMs handle visual question ambiguities in real-world scenes through the interactive clarification approach emphasized in our work. These examples are generated by $\mathcal{L}\text{LaVA-13B} \text{(SFT+DPO)}$, which performs best in our experiments.

\end{document}